\documentclass[]{interact}

\usepackage{epstopdf}
\usepackage{subfigure}
\usepackage{tabularx}
\usepackage{graphicx}
\usepackage{amsmath}
\usepackage{amssymb}
\usepackage{booktabs}
\usepackage{placeins} 
\usepackage{float} 
\usepackage{enumitem}
\usepackage{tabularx} 
\usepackage{longtable} 
\usepackage{array} 
\usepackage{ragged2e} 
\usepackage{booktabs} 
\usepackage{placeins}
\usepackage{afterpage} 
\usepackage{caption} 
\usepackage{booktabs} 
\usepackage{adjustbox} 
\usepackage{graphicx} 
\usepackage{tabularx}

\theoremstyle{plain}

\theoremstyle{definition}

\theoremstyle{remark}

\begin{document}


\title{Hyperspectral Image Classification via Transformer-based Spectral-Spatial Attention Decoupling and Adaptive Gating}

\author{
\name{Guandong Li\textsuperscript{a}\thanks{CONTACT Guandong Li. Email: leeguandon@gmail.com} and Mengxia Ye\textsuperscript{b}}
\affil{\textsuperscript{a}iFLYTEK, Shushan, Hefei, Anhui, China; \textsuperscript{b}Aegon THTF,Qinghuai,Nanjing,Jiangsu,China}
}

\maketitle

\begin{abstract}
Deep neural networks face several challenges in hyperspectral image classification, including high-dimensional data, sparse distribution of ground objects, and spectral redundancy, which often lead to classification overfitting and limited generalization capability. To more effectively extract and fuse spatial context with fine spectral information in hyperspectral image (HSI) classification, this paper proposes a novel network architecture called STNet. The core advantage of STNet stems from the dual innovative design of its Spatial-Spectral Transformer module: first, the fundamental explicit decoupling of spatial and spectral attention ensures targeted capture of key information in HSI; second, two functionally distinct gating mechanisms perform intelligent regulation at both the fusion level of attention flows (adaptive attention fusion gating) and the internal level of feature transformation (GFFN). This characteristic demonstrates superior feature extraction and fusion capabilities compared to traditional convolutional neural networks, while reducing overfitting risks in small-sample and high-noise scenarios. STNet enhances model representation capability without increasing network depth or width. The proposed method demonstrates superior performance on IN, UP, and KSC datasets, outperforming mainstream hyperspectral image classification approaches.
\end{abstract}


\section{Introduction}

Hyperspectral remote sensing images (HSI) play a crucial role in spatial information applications due to their unique narrow-band imaging characteristics. The imaging equipment synchronously records both spectral and spatial position information of sampling points, integrating them into a three-dimensional data structure containing two-dimensional space and one-dimensional spectrum. As an important application of remote sensing technology, ground object classification demonstrates broad value in fields including ecological assessment, transportation planning, agricultural monitoring, land management, and geological surveys \cite{chang2003hyperspectral,bing2011intelligent}. However, hyperspectral remote sensing images face several challenges in ground object classification. First, hyperspectral data typically exhibits high-dimensional characteristics, with each pixel containing hundreds or even thousands of bands, leading to data redundancy and computational complexity while potentially causing the "curse of dimensionality," making classification models prone to overfitting under sparse sample conditions. Second, sparse ground object distribution means training samples are often limited, particularly for certain rare categories where annotation costs are high and distributions are imbalanced, further constraining model generalization capability. Additionally, hyperspectral images are frequently affected by noise, atmospheric interference, and mixed pixels, reducing signal-to-noise ratio and increasing difficulty in extracting features from sparse ground objects. Finally, ground object sparsity may also lead to insufficient spatial context information, causing loss in classification and recognition.

Deep learning methods for HSI classification \cite{li2019doubleconvpool,li2020hyperspectral,li2022faster,li2023dgcnet,li2025spatial_geometry,li20253d,li2018scene} have achieved significant progress. In \cite{lee2017going} and \cite{zhao2016spectral}, Principal Component Analysis (PCA) was first applied to reduce the dimensionality of the entire hyperspectral data, followed by extracting spatial information from neighboring regions using 2D CNN. Methods like 2D-CNN \cite{makantasis2015deep,chen2016deep} require separate extraction of spatial and spectral features, failing to fully utilize joint spatial-spectral information and necessitating complex preprocessing. \cite{wang2018fast} proposed a Fast Dense Spectral-Spatial Convolutional Network (FDSSC) based on dense networks, constructing 1D-CNN and 3D-CNN dense blocks connected in series. FSKNet \cite{li2022faster} introduced a 3D-to-2D module and selective kernel mechanism, while 3D-SE-DenseNet \cite{li2020hyperspectral} incorporated the SE mechanism into 3D-CNN to correlate feature maps between different channels, activating effective information while suppressing ineffective information in feature maps. DGCNet \cite{li2023dgcnet} designed dynamic grouped convolution (DGC) on 3D convolution kernels, where DGC introduces small feature selectors for each group to dynamically determine which part of input channels to connect based on activations of all input channels. Multiple groups can capture different complementary visual/semantic features of input images, enabling CNNs to learn rich feature representations. DHSNet \cite{liu2025dual} proposed a novel Central Feature Attention-Aware Convolution (CFAAC) module that guides attention to focus on central features crucial for capturing cross-scene invariant information. To leverage the advantages of both CNN and Transformer, many studies have attempted to combine CNN and Transformer to utilize local and global feature information of HSI. \cite{sun2022spectral} proposed a Spectral-Spatial Feature Tokenization Transformer (SSFTT) network that extracts shallow features through 3D and 2D convolutional layers and uses Gaussian-weighted feature tokens to extract high-level semantic features in the transformer encoder. In \cite{liu2024kan}, the authors demonstrated that KANs significantly outperform traditional multilayer perceptrons (MLPs) in satellite traffic prediction tasks, achieving more accurate predictions with fewer learnable parameters. In \cite{vaca2024kolmogorov}, KANs were recognized as a promising alternative for efficient image analysis in remote sensing, highlighting their effectiveness in this field.Some Transformer-based methods \cite{hong2021spectralformer,zhao2016spectral,liu2021swin} employ grouped spectral embedding and transformer encoder modules to model spectral representations, but these methods have obvious shortcomings - they treat spectral bands or spatial patches as tokens and encode all tokens, resulting in significant redundant computations. However, HSI data already contains substantial redundant information, and their accuracy often falls short compared to 3D-CNN-based methods while requiring greater computational complexity.

3D-CNN possesses the capability to sample simultaneously in both spatial and spectral dimensions, maintaining the spatial feature extraction ability of 2D convolution while ensuring effective spectral feature extraction. 3D-CNN can directly process high-dimensional data, eliminating the need for preliminary dimensionality reduction of hyperspectral images. However, the 3D-CNN paradigm has significant limitations - when simultaneously extracting spatial and spectral features, it may incorporate irrelevant or inefficient spatial-spectral combinations into computations. For instance, certain spatial features may be prominent in specific bands while appearing as noise or irrelevant information in other bands, yet 3D convolution still forcibly combines these low-value features. Computational and dimensional redundancy can easily trigger overfitting risks, further limiting model generalization capability. Currently widely used methods such as DFAN \cite{zhang2020deep}, MSDN \cite{zhang2019multi}, 3D-DenseNet \cite{zhang2019three}, and 3D-SE-DenseNet\cite{li2020hyperspectral} employ operations like dense connections. While dense connections directly link each layer to all its preceding layers, enabling feature reuse, they introduce redundancy when subsequent layers do not require early features. Therefore, how to more efficiently enhance the representational capability of 3D convolution kernels in 3D convolution, achieving more effective feature extraction with fewer cascaded 3D convolution kernels and dense connections while optimizing the screening and skipping of redundant information has become a direction in hyperspectral classification.

This work addresses the challenge of effectively extracting and fusing spatial context with fine spectral information in hyperspectral image (HSI) classification by proposing a novel network architecture called STNet. The core of STNet lies in its SpatioTemporalTransformer module, whose primary innovation is explicitly decoupling the attention mechanism into two parallel and focused branches: spatial attention (spatial\_attn) and spectral attention (temporal\_attn). We recognize that HSI data simultaneously contains rich spatial structures and unique spectral sequence information, which traditional methods often struggle to capture in a balanced and efficient manner. Therefore, STNet specifically designs the spatial attention branch to focus on capturing pixel neighborhood spatial structures and texture features within each spectral band (Intra-band Spatial Dependencies); meanwhile, the spectral attention branch concentrates on learning feature responses and long-range correlations between different spectral bands (Inter-band Spectral Correlations) through aggregation of spatial dimensions (such as mean pooling). This explicit division of responsibilities forms the foundation for STNet's effective processing of HSI data, ensuring targeted extraction of these two types of key information. Building upon this dual-branch attention framework, STNet introduces its first key gating mechanism: the Adaptive Attention Fusion Gate (implemented in gate). This gating unit is positioned after the outputs of both spatial and spectral attention branches, serving as an intelligent coordinator. It learns a dynamic scalar weight in a data-driven manner to intelligently balance and weight the fusion of information flows from these two independent attention branches. This adaptive weighting strategy enables the model to dynamically adjust its integration focus based on specific characteristics of input data—for example, when spectral information is more critical in certain regions while spatial structures are more important in others—achieving synergistic integration of spectral-spatial information and significantly enhancing representation capability for complex ground object scenes. This gating mechanism specifically addresses the problem of optimally combining these two pre-separated, different-modal attention information sources. Furthermore, in the feature transformation stage, STNet employs a second distinct type of gating mechanism: feature gating within the feed-forward network (Gated Feed-Forward Network, GFFN). The standard Transformer FFN is responsible for nonlinear transformation of attention-integrated features. To refine feature representation at this stage, STNet introduces GFFN within the FFN. Unlike the aforementioned fusion gate, GFFN operates within the FFN module itself. It runs parallel to the main FFN path, learning an element-wise gating vector with the same feature dimensions as the input, which is then applied to the main FFN output through element-wise multiplication. This mechanism grants the model finer-grained internal control over information flow, enabling dynamic and selective modulation of FFN output feature channels to enhance important features while suppressing redundant information. This gating mechanism focuses on solving the problem of how to perform more effective feature selection and filtering during nonlinear transformation, operating within already-fused feature flows. In summary, the core advantages of STNet stem from the dual innovative design of its SpatioTemporalTransformer module: first, the fundamental explicit decoupling of spatial and spectral attention ensures targeted capture of key HSI information; second, two functionally distinct gating mechanisms perform intelligent regulation at both the fusion level of attention flows (adaptive attention fusion gating) and the internal level of feature transformation (GFFN). The former is responsible for intelligently combining heterogeneous attention sources, while the latter refines internal representation within single information flows. This structure not only enhances representation of key spatial-spectral features through decoupled attention but also, compared to traditional 3D-CNN, can more effectively focus on high-value information while reducing redundant interference, avoiding uniform treatment of all spectral dimensions, thereby effectively alleviating parameter redundancy and the curse of dimensionality, ultimately improving classification accuracy.

The main contributions of this paper are as follows:

1. This paper proposes a novel STNet architecture that introduces a spatial-spectral Transformer model, improving the efficient 3D-DenseNet-based joint spatial-spectral hyperspectral image classification method. It includes feature extraction through spatial-spectral Transformer and intelligent feature extraction and distribution through gating mechanisms, addressing overfitting risks caused by spatial information redundancy and the curse of dimensionality in hyperspectral data while enhancing network generalization capability. By combining dense connections with dynamic convolution generation—where dense connections facilitate feature reuse in the network—the designed 3D-DenseNet model achieves good accuracy on both IN and UP datasets.

2. This paper introduces a Transformer module in the 3D-CNN structure, proposing a novel SpatioTemporalTransformer module that achieves targeted and independent modeling of spatial context information and spectral correlations in hyperspectral images (HSI) through explicit decoupling of spatial and spectral attention mechanisms. Additionally, it introduces two functionally complementary gating mechanisms: the Adaptive Attention Fusion Gate and the Gated Feed-Forward Network (GFFN). The former is responsible for intelligently combining heterogeneous attention sources, while the latter refines internal representation within single information flows.

3. STNet is more concise than networks combining various DL mechanisms, without complex connections and concatenations, requiring less computation. Without increasing network depth or width, it improves model representation capability through wavelet convolution with expanded receptive fields.

\section{SpatioTemporal Transformer with Adaptive Gating for Classification}
\subsection{Novel Structure Design in 3D-CNN}

Given the sparse and finely clustered characteristics of hyperspectral ground objects, most methods consider how to enhance the model's feature extraction capability regarding spatial-spectral dimensions, making novel structure design an excellent approach to strengthen spatial-spectral dimension information representation. LGCNet\cite{li2025spatial} designed a learnable grouped convolution structure where both input channels and convolution kernel groups can be learned end-to-end through the network. DGCNet\cite{li2023dgcnet} designed dynamic grouped convolution, introducing small feature selectors for each group to dynamically determine which part of input channels to connect based on activations of all input channels. DACNet\cite{li2025efficient} designed dynamic attention convolution using SE to generate weights and multiple parallel convolutional kernels instead of single convolution. SG-DSCNet\cite{li2025spatial_geometry} designed a Spatial-Geometry Enhanced 3D Dynamic Snake Convolutional Neural Network, introducing deformable offsets in 3D convolution to increase kernel flexibility through constrained self-learning processes, thereby enhancing the network's regional perception of ground objects and proposing multiview feature fusion. WCNet\cite{li20253d} designed a 3D convolutional network integrating wavelet transform, introducing wavelet transform to expand the receptive field of convolution through wavelet convolution, guiding CNN to better respond to low frequencies through cascading. Each convolution focuses on different frequency bands of the input signal with gradually increasing scope. EKGNet\cite{li2025expert} includes a context-associated mapping network and dynamic kernel generation module, where the context-associated mapping module translates global contextual information of hyperspectral inputs into instructions for combining base convolutional kernels, while dynamic kernels are composed of K groups of base convolutions, analogous to K different types of experts specializing in fundamental patterns across various dimensions. These methods are all based on the design concept of adaptive convolution, attempting to address the complexity of hyperspectral data in spatial-spectral dimensions through adaptive convolution mechanisms. Unlike these methods that focus on improving the convolution operation itself or introducing dynamic mechanisms, the STNet proposed in this paper adopts another strategy: introducing a carefully designed SpatioTemporalTransformer module within the 3D-CNN framework. This module aims to more effectively combine 3D-CNN's local feature extraction capability with Transformer's advantages in global dependency modeling and adaptive information integration through explicit spatial-spectral attention decoupling and dual gating mechanisms (adaptive fusion gating and gated feed-forward network) to address the unique challenges of HSI classification.

\begin{figure}[h]
\centering
\includegraphics[width=0.9\linewidth]{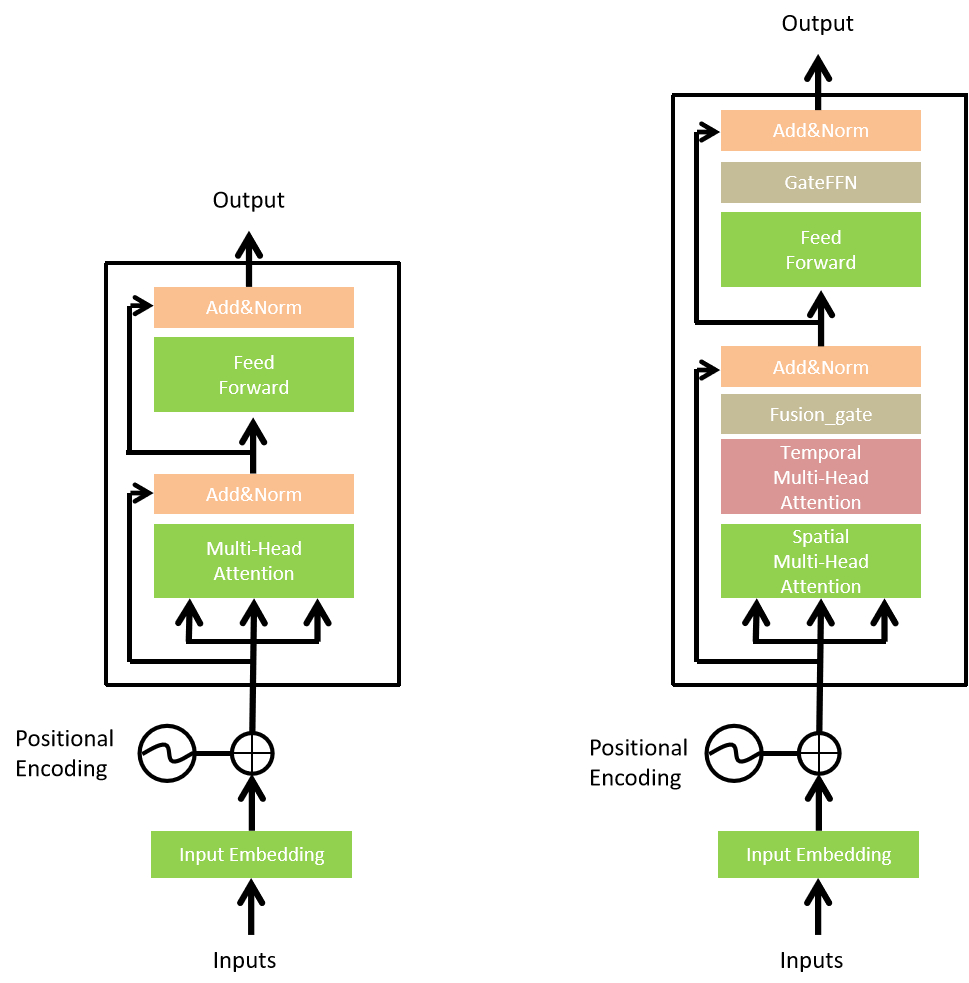}
\caption{Left: Standard self-attention structure; Right: Our SpatioTemporalTransformer structure}
\label{fig:transformer_structure}
\end{figure}

\subsection{SpatioTemporal Transformer with Adaptive Gating Mechanism}

Hyperspectral image sample data is scarce and exhibits sparse ground object characteristics, with uneven spatial distribution and substantial redundant information in the spectral dimension. Although 3D-CNN structures can utilize joint spatial-spectral information, how to more effectively achieve deep extraction of spatial-spectral information remains a noteworthy issue. As the core of convolutional neural networks, convolution kernels are generally regarded as information aggregators that combine spatial information and feature dimension information in local receptive fields. Convolutional neural networks consist of a series of convolutional layers, nonlinear layers, and downsampling layers, enabling them to capture image features from a global receptive field for image description. However, training a high-performance network is challenging, and much work has been done to improve network performance from the spatial dimension perspective. For example, the Residual structure achieves deep network extraction by fusing features produced by different blocks, while DenseNet enhances feature reuse through dense connections. 3D-CNN contains numerous redundant weights in feature extraction through convolution operations that simultaneously process spatial and spectral information of hyperspectral images. This redundancy is particularly prominent in joint spatial-spectral feature extraction: from the spatial dimension, ground objects in hyperspectral images are sparsely and unevenly distributed, and convolutional kernels may capture many irrelevant or low-information regions within local receptive fields; from the spectral dimension, hyperspectral data typically contains hundreds of bands with high correlation and redundancy between adjacent bands, making weight allocation of convolutional kernels along the spectral axis difficult to effectively focus on key features. Especially the redundant characteristics of spectral dimensions cause many convolutional parameters to only serve as "fillers" in high-dimensional data without fully mining deep patterns in joint spatial-spectral information. This weight redundancy not only increases computational complexity but may also weaken the model's representation capability for sparse ground objects and complex spectral features, thereby limiting 3D-CNN's performance in hyperspectral image processing.

STNet proposes a novel solution, with its core lying in the SpatioTemporalTransformer module embedded within the 3D-CNN architecture (such as the DenseNet structure used in this study). This module first explicitly decouples the attention mechanism into independent spatial attention (spatial\_attn) and spectral attention (temporal\_attn) branches to specifically capture complex local spatial structures and long-range spectral dependencies in HSI data. Subsequently, it uses an adaptive attention fusion gate (gate) to dynamically learn weights based on data to intelligently balance and weight the fusion of these two heterogeneous information flows. Furthermore, during the feature transformation stage, it employs a gated feed-forward network (GFFN, gate\_ffn) to perform refined selection and enhancement of fused features. This design aims to directly address the problems faced by traditional 3D-CNN when dealing with high-dimensional redundancy and sparse ground object distribution: inefficient utilization of convolutional weights, difficulty in effectively focusing on key spatial-spectral features, and potential forced fusion of low-value information. It achieves more efficient, more adaptive deep extraction and representation of joint spatial-spectral features to improve performance and generalization capability in hyperspectral image classification tasks. Compared with traditional three-dimensional convolution, this module not only preserves the spatial locality of convolution operations but also achieves higher parameter utilization efficiency. Experimental verification shows it achieves excellent classification performance in hyperspectral classification tasks with fewer parameters. In hyperspectral applications, this module demonstrates strong adaptability through fine modeling of complex ground object spectral features. Combined with the deep characteristics of 3D-DenseNet, it can extract features more effectively. This marks an important development direction in hyperspectral data processing: moving from reliance on fixed, homogeneous convolutional filters to adopting more flexible, attention-driven hybrid architectures with adaptive information integration capabilities to address the challenges of high-dimensional complex data.

\begin{figure}[h]
\centering
\includegraphics[width=0.9\linewidth]{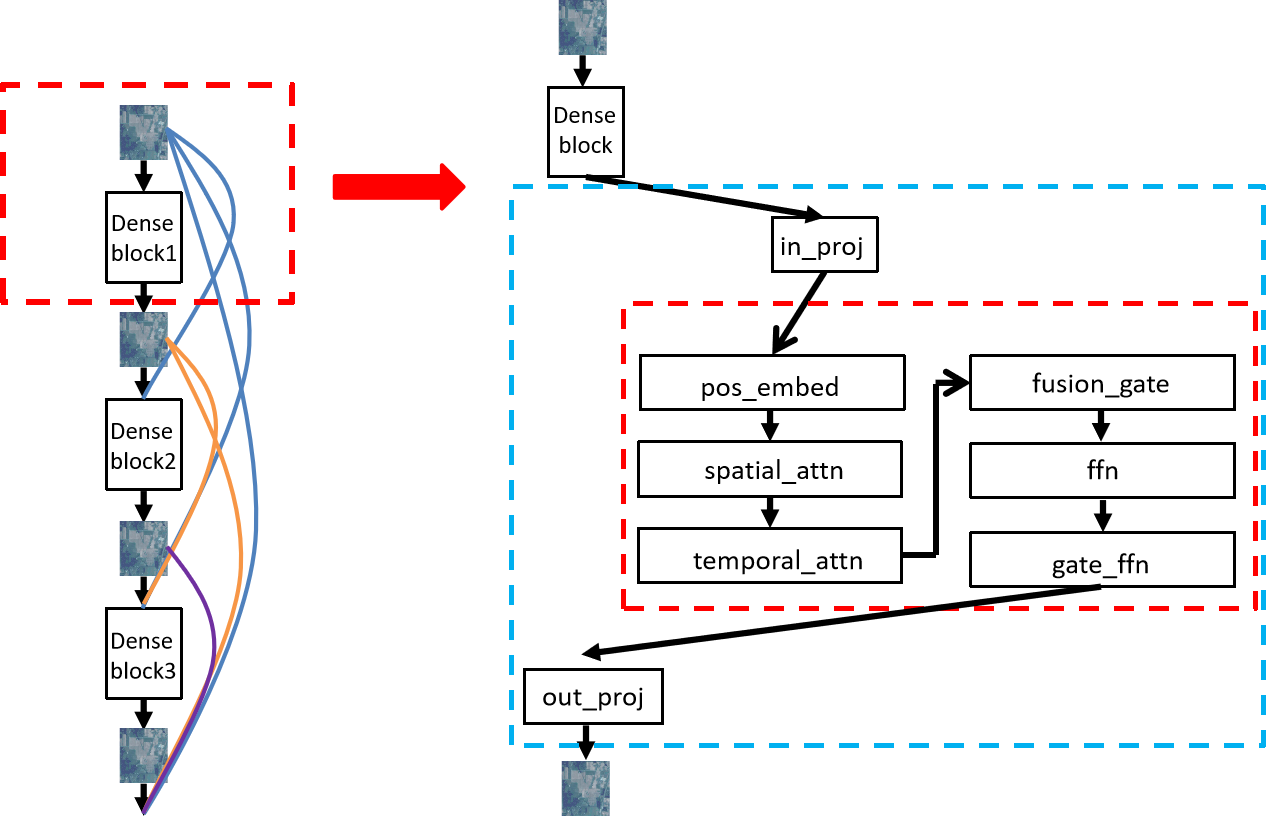}
\caption{Design of SpatioTemporalTransformer in 3D-DenseNet's dense block}
\label{fig:dense_block_design}
\end{figure}

\subsubsection{Explicit Spectral-Spatial Attention Decoupling with Adaptive Fusion}

The hyperspectral image (HSI) data cube $\mathbf{X} \in \mathbb{R}^{B \times D \times H \times W \times C_{in}}$ (where $B$ is batch size, $D$ is number of spectral bands, $H$, $W$ are spatial dimensions, and $C_{in}$ is initial number of channels) simultaneously contains rich spatial context information and fine spectral sequence features. Traditional methods often face challenges in jointly extracting this information. To achieve more precise and efficient feature extraction, the SpatioTemporalTransformer module first maps $\mathbf{X}$ to feature space $\mathbf{X}' \in \mathbb{R}^{B \times D \times H \times W \times C}$ (where $C$ is $d_{model}$) through the input projection layer in\_proj, then introduces an explicit spectral-spatial attention decoupling mechanism.

1. Spatial attention (spatial\_attn): This branch focuses on capturing spatial dependency relationships within each spectral band. We reshape $\mathbf{X}'$ into $\mathbf{X}_s \in \mathbb{R}^{(H \times W) \times (B \times D) \times C}$, flattening the spatial dimension as sequence length while merging batch and spectral dimensions. Then we apply multi-head self-attention (MHSA):

\begin{equation}
\mathbf{Q}_s, \mathbf{K}_s, \mathbf{V}_s = \mathbf{X}_s \mathbf{W}^Q_s, \mathbf{X}_s \mathbf{W}^K_s, \mathbf{X}_s \mathbf{W}^V_s
\end{equation}

\begin{equation}
\text{AttnOutput}_s = \text{MHSA}_{\text{spatial}}(\mathbf{Q}_s, \mathbf{K}_s, \mathbf{V}_s) \in \mathbb{R}^{(H \times W) \times (B \times D) \times C}
\end{equation}

where $\mathbf{W}^Q_s$, $\mathbf{W}^K_s$, $\mathbf{W}^V_s$ are learnable projection matrices for the spatial attention branch. This process simultaneously calculates inter-pixel spatial relationships for all $D$ bands, effectively capturing local textures and spatial structures (Intra-band Spatial Dependencies).

2. Spectral attention (temporal\_attn): This branch focuses on modeling long-range correlations between different spectral bands. First, we perform average pooling on the spatial dimensions of $\mathbf{X}'$ to obtain representative features for each band, then reshape into $\mathbf{X}_t \in \mathbb{R}^{D \times B \times C}$:

\begin{equation}
\mathbf{X}_{\text{pooled}} = \text{MeanPool}_{H,W}(\mathbf{X}') \in \mathbb{R}^{B \times D \times 1 \times 1 \times C}
\end{equation}

\begin{equation}
\mathbf{X}_t = \text{Reshape}(\mathbf{X}_{\text{pooled}}) \in \mathbb{R}^{D \times B \times C}
\end{equation}

Subsequently, treating spectral dimension $D$ as sequence length, we apply multi-head self-attention:

\begin{equation}
\mathbf{Q}_t, \mathbf{K}_t, \mathbf{V}_t = \mathbf{X}_t \mathbf{W}^Q_t, \mathbf{X}_t \mathbf{W}^K_t, \mathbf{X}_t \mathbf{W}^V_t
\end{equation}

\begin{equation}
\text{AttnOutput}_t = \text{MHSA}_{\text{temporal}}(\mathbf{Q}_t, \mathbf{K}_t, \mathbf{V}_t) \in \mathbb{R}^{D \times B \times C}
\end{equation}

where $\mathbf{W}^Q_t$, $\mathbf{W}^K_t$, $\mathbf{W}^V_t$ are projection matrices for the spectral attention branch. This helps identify key spectral features and their cross-band patterns (Inter-band Spectral Correlations).

To intelligently integrate these two different sources of attention information, we designed an adaptive attention fusion gate (gate). This gate learns a dynamic weight $g$ to balance the contributions of $\text{AttnOutput}_s$ and $\text{AttnOutput}_t$. We first aggregate the two attention outputs to obtain representative vectors (possibly using mean operation in the code), denoted as $h_s$ and $h_t$. The gate weight $g$ is calculated through a small feed-forward network (self.gate, containing linear layers Linear\_g1, Linear\_g2 and activation functions ReLU, Sigmoid $\sigma$):

\begin{equation}
\text{gate\_input} = \text{Concat}(h_s, h_t)
\end{equation}

\begin{equation}
g = \sigma(\text{Linear}_{g2}(\text{ReLU}(\text{Linear}_{g1}(\text{gate\_input}))))
\end{equation}

The final fused attention output $\text{AttnOutput}_{\text{fused}}$ (before being added to input $\mathbf{x}_{\text{flat}}$) is obtained through weighted summation, where $\text{AttnOutput}'_t$ is the result of $\text{AttnOutput}_t$ after appropriate shape adjustment to match $\text{AttnOutput}_s$:

\begin{equation}
\text{AttnOutput}_{\text{fused}} = g \ast \text{AttnOutput}_s + (1 - g) \ast \text{AttnOutput}'_t
\end{equation}

Here, the gate weight $g$ (which may be a scalar or a channel-related vector, broadcast according to code implementation) allows the model to dynamically adjust fusion strategies based on specific spatial-spectral characteristics of input HSI data, achieving optimal spatial-spectral information synergy.

\subsubsection{Learnable and Interpolatable 3D Positional Encoding}

The Transformer architecture lacks inherent perception of sequence order and requires positional encoding (PE) to inject positional information. For HSI cube data $\mathbf{X}' \in \mathbb{R}^{B \times D \times H \times W \times C}$, its three-dimensional (spectral $D$, spatial $H$, $W$) coordinate information is crucial. We abandoned fixed or size-limited positional encoding schemes and adopted parameterized, learnable 3D positional encoding (pos\_embed). This encoding is initialized as a learnable tensor $\mathbf{P}_{\text{learn}} \in \mathbb{R}^{1 \times C \times D_{\text{init}} \times H_{\text{init}} \times W_{\text{init}}}$, with its values optimized during training through backpropagation, enabling the model to autonomously learn position representations most suitable for HSI data characteristics and downstream classification tasks.

To accommodate possible variations in input block sizes ($D$, $H$, $W$) in HSI analysis, we introduced a dynamic size adaptation mechanism based on trilinear interpolation. During each forward pass, the pre-learned positional encoding $\mathbf{P}_{\text{learn}}$ is adjusted through interpolation according to current input $\mathbf{X}'$'s actual dimensions ($D$, $H$, $W$):

\begin{equation}
\mathbf{P}_{\text{adjusted}} = \text{Interpolate}_{\text{Trilinear}}(\mathbf{P}_{\text{learn}}, \text{size}=(D, H, W)) \in \mathbb{R}^{1 \times C \times D \times H \times W}
\end{equation}

where $\text{Interpolate}_{\text{Trilinear}}$ represents the torch.nn.functional.interpolate function using 'trilinear' mode. The interpolated positional encoding $\mathbf{P}_{\text{adjusted}}$ is then appropriately reshaped (Reshape) and added to the flattened input features $\mathbf{X}'_{\text{flat}} \in \mathbb{R}^{(B \times D \times H \times W) \times C}$ (assuming $\mathbf{X}'$ has been flattened):

\begin{equation}
\mathbf{X}'_{\text{positioned}} = \mathbf{X}'_{\text{flat}} + \text{Reshape}(\mathbf{P}_{\text{adjusted}})
\end{equation}

(Note: The addition operation requires ensuring $\text{Reshape}(\mathbf{P}_{\text{adjusted}})$'s shape is compatible with $\mathbf{X}'_{\text{flat}}$, possibly involving broadcasting or repetition). This learnable and interpolatable design not only makes position information representation more aligned with data characteristics but also endows the model with flexibility and robustness in processing inputs of different sizes, eliminating the need to retrain or design specific networks for different input dimensions.

\subsubsection{Gated Feed-Forward Network (GFFN)}
In the SpatioTemporalTransformer module, the feature representation after attention fusion and the first residual connection and layer normalization (Layer Normalization, LN), denoted as $\mathbf{Y} \in \mathbb{R}^{N \times C}$ (where $N = B \times D \times H \times W$ is sequence length, $C$ is $d_{model}$), is input into the gated feed-forward network (GFFN) for nonlinear feature transformation. Standard FFN typically contains two linear layers and activation functions. To enhance feature selectivity and information flow control, we introduced a gating mechanism. GFFN contains two parallel paths:

1. Main data path (self.ffn): Performs primary nonlinear transformation on input $\mathbf{Y}$, typically containing two linear layers (Linear\_ffn1, Linear\_ffn2) and an activation function (such as GELU). Ignoring Dropout, its calculation can be expressed as:

\begin{equation}
\mathbf{H}_{\text{main}} = \text{Linear}_{\text{ffn2}}(\text{GELU}(\text{Linear}_{\text{ffn1}}(\mathbf{Y}))) \in \mathbb{R}^{N \times C}
\end{equation}

2. Gating path (self.gate\_ffn): In parallel, input $\mathbf{Y}$ passes through an independent linear layer (Linear\_gate) and a Sigmoid activation function $\sigma$, generating a gating vector $\mathbf{g}_{\text{ffn}}$ with the same dimensions as the main path output:

\begin{equation}
\mathbf{g}_{\text{ffn}} = \sigma(\text{Linear}_{\text{gate}}(\mathbf{Y})) \in \mathbb{R}^{N \times C}
\end{equation}

Each element value of this gating vector $\mathbf{g}_{\text{ffn}}$ falls within the (0, 1) interval.

The final GFFN output $\text{Output}_{\text{GFFN}}$ is obtained by element-wise multiplication (Hadamard product, $\odot$) of the main path output $\mathbf{H}_{\text{main}}$ and the gating vector $\mathbf{g}_{\text{ffn}}$:

\begin{equation}
\text{Output}_{\text{GFFN}} = \mathbf{H}_{\text{main}} \odot \mathbf{g}_{\text{ffn}} \in \mathbb{R}^{N \times C}
\end{equation}

This gating mechanism allows the network to dynamically modulate each feature dimension in $\mathbf{H}_{\text{main}}$. When gating values $\mathbf{g}_{\text{ffn}}$ approach 1, corresponding features are preserved; when approaching 0, they are suppressed. This enables GFFN to adaptively enhance important features while weakening redundant or noisy information, achieving more refined feature selection and filtering. In hyperspectral image classification, facing subtle spectral differences and complex spatial patterns, this enhanced feature transformation capability helps the model learn more discriminative deep representations, thereby improving classification performance.

This explicit spectral-spatial attention decoupling design enables STNet to efficiently capture subtle changes in the spectral dimension, particularly its spectral attention branch (temporal\_attn) focusing on long-range spectral correlations, while the adaptive gating mechanism (fusion gate and GFFN gate\_ffn) significantly reduces overfitting risks in small-sample or high-noise HSI scenarios through intelligent screening and weighting of information flows. Compared with traditional three-dimensional convolution, STNet's SpatioTemporalTransformer module preserves the spatial locality provided by the underlying 3D-CNN architecture (such as 3D-DenseNet) while achieving dynamic, refined control over feature representation through its attention mechanism and gating. Experimental verification shows it achieves excellent classification performance in hyperspectral classification tasks, demonstrating the potential to enhance model expression capability and generalization through structural innovation. By integrating the SpatioTemporalTransformer module into the 3D-DenseNet framework, STNet benefits from enhanced feature reuse while its unique attention decoupling and gating mechanisms further optimize feature representation, effectively addressing redundancy issues that may arise from dense connections. The synergy between decoupled attention, adaptive fusion, learnable position encoding, and gated feed-forward networks ensures STNet can effectively alleviate challenges posed by high-dimensional redundancy and sparse ground object distribution, achieving exceptional classification performance.

\begin{figure}[h]
\centering
\includegraphics[width=0.9\linewidth]{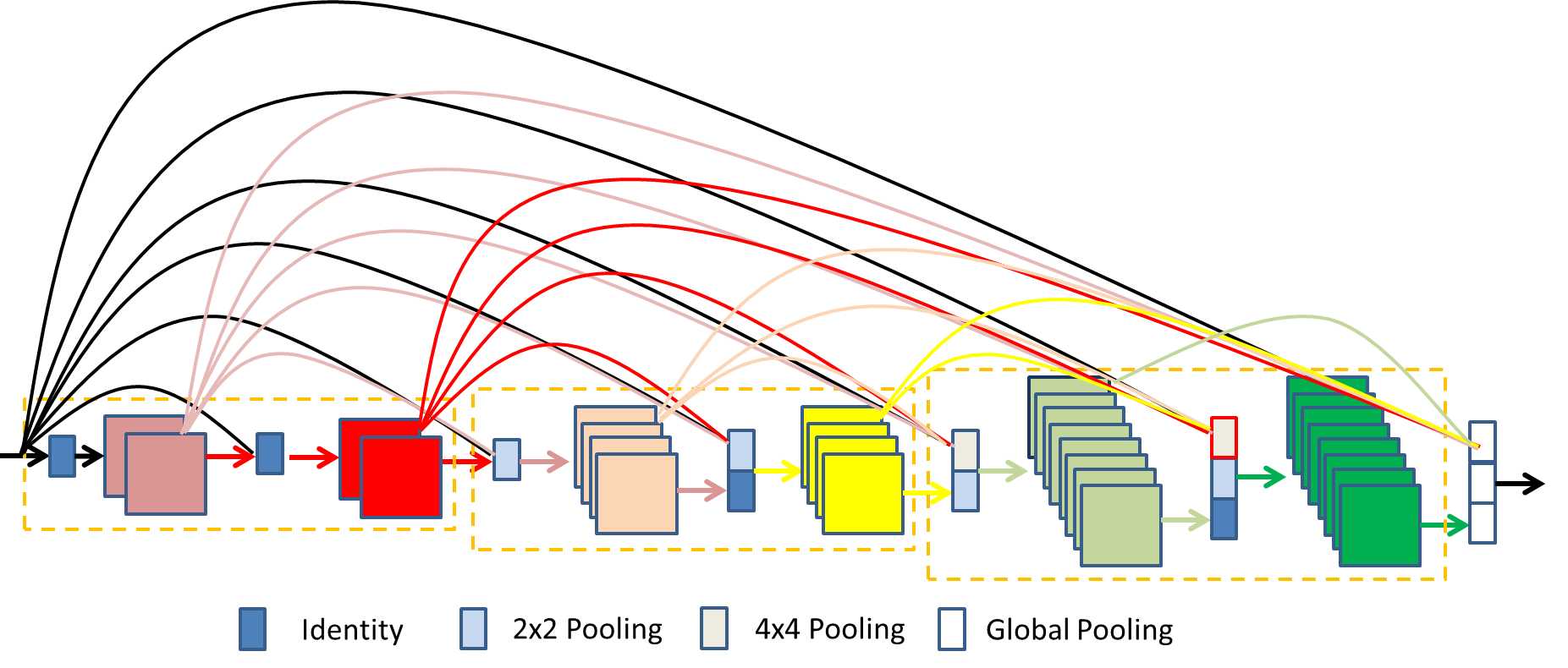}
\caption{Proposed DenseNet variant with two key differences from original DenseNet: (1) Direct connections between layers with different feature resolutions; (2) Growth rate doubles when feature map size reduces (third yellow dense block generates significantly more features than the first block)}
\label{fig:densenet_variant}
\end{figure}

\subsection{3D-CNN Framework for Hyperspectral Image Feature Extraction and Model Implementation}
We made two modifications to the original 3D-DenseNet to further simplify the architecture and improve its computational efficiency.

\subsubsection{Exponentially Increasing Growth Rate}
The original DenseNet design adds $k$ new feature maps per layer, where $k$ is a constant called the growth rate. As shown in \cite{huang2017densely}, deeper layers in DenseNet tend to rely more on high-level features than low-level ones, which motivated our improvement through strengthened short connections. We found this could be achieved by gradually increasing the growth rate with depth. This increases the proportion of features from later layers relative to earlier ones. For simplicity, we set the growth rate as $k=2^{m-1}k_0$, where $m$ is the dense block index and $k_0$ is a constant. This growth rate setting does not introduce any additional hyperparameters. The "increasing growth rate" strategy places a larger proportion of parameters in the model's later layers. This significantly improves computational efficiency, though it may reduce parameter efficiency in some cases. Depending on specific hardware constraints, trading one for the other may be advantageous \cite{liu2017learning}.

\subsubsection{Fully Dense Connectivity}
To encourage greater feature reuse than the original DenseNet architecture, we connect the input layer to all subsequent layers across different dense blocks (see Figure~\ref{fig:densenet_variant}). Since dense blocks have different feature resolutions, we downsample higher-resolution feature maps using average pooling when connecting them to lower-resolution layers.

The overall model architecture is shown below:

\begin{figure}[h]
\centering
\includegraphics[width=0.9\linewidth]{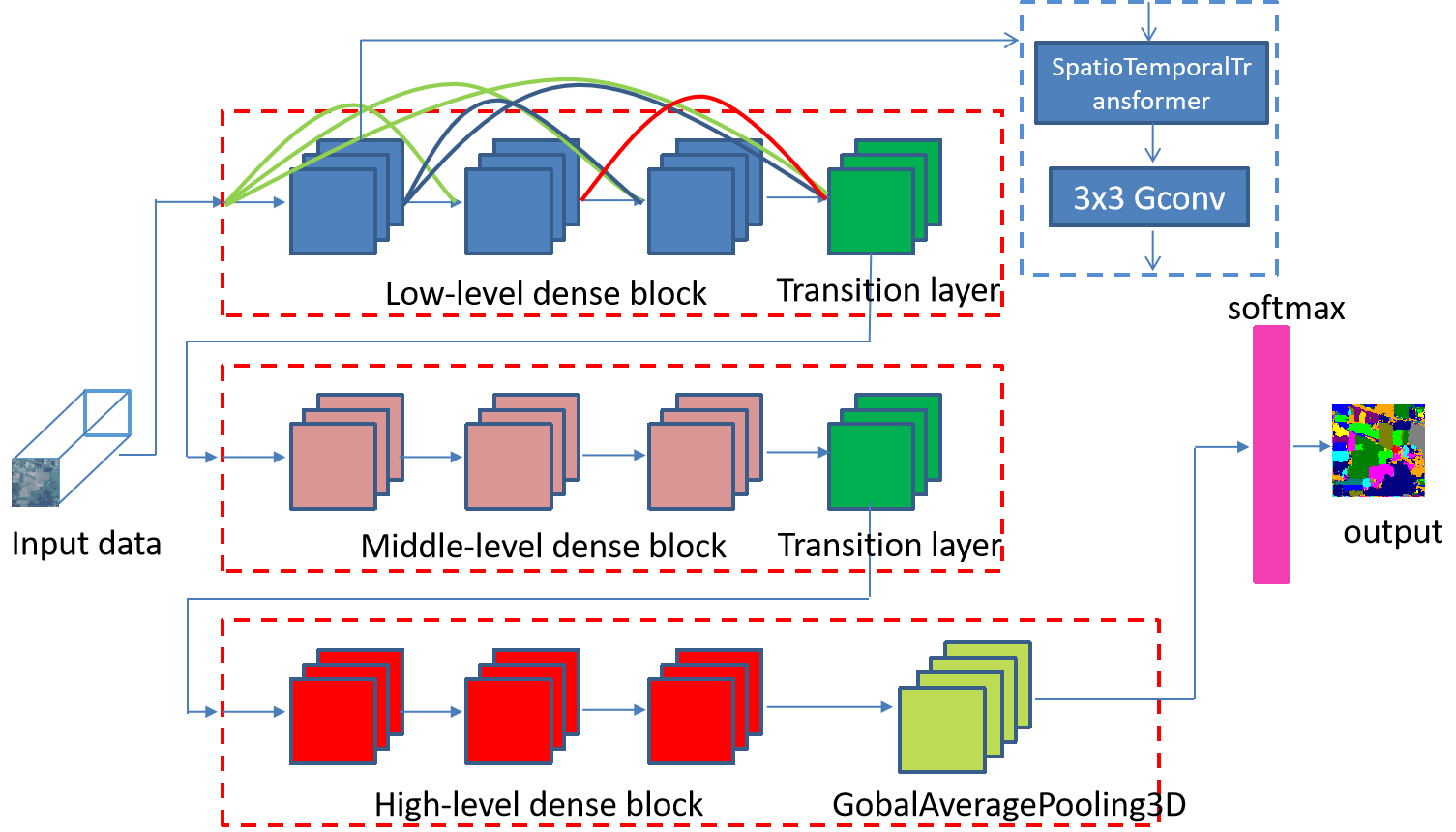}
\caption{Overall architecture of our STNet, incorporating the 3D-DenseNet basic framework}
\label{fig:stnet_architecture}
\end{figure}

\section{Experiments and Analysis}

To evaluate the performance of STNet, we conducted experiments on three representative hyperspectral datasets: Indian Pines, Pavia University, and Kennedy Space Center (KSC). The classification metrics include Overall Accuracy (OA), Average Accuracy (AA), and Kappa coefficient.

\subsection{Experimental Datasets}
\subsubsection{Indian Pines Dataset}
\begin{figure}[h]
\centering
\includegraphics[width=0.9\linewidth]{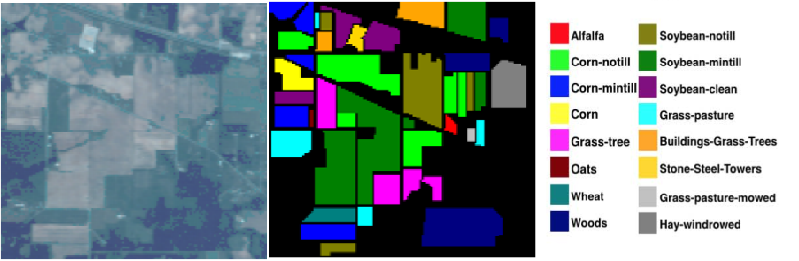}
\caption{False color composite and ground truth labels of Indian Pines dataset}
\label{fig:indian_pines}
\end{figure}

The Indian Pines dataset was collected in June 1992 by the AVIRIS (Airborne Visible/Infrared Imaging Spectrometer) sensor over a pine forest test site in northwestern Indiana, USA. The dataset consists of $145\times145$ pixel images with a spatial resolution of 20 meters, containing 220 spectral bands covering the wavelength range of 0.4--2.5$\mu$m. In our experiments, we excluded 20 bands affected by water vapor absorption and low signal-to-noise ratio (SNR), utilizing the remaining 200 bands for analysis. The dataset encompasses 16 land cover categories including grasslands, buildings, and various crop types. Figure~\ref{fig:indian_pines} displays the false-color composite image and spatial distribution of ground truth samples.

\subsubsection{Pavia University Dataset}
\begin{figure}[h]
\centering
\includegraphics[width=0.9\linewidth]{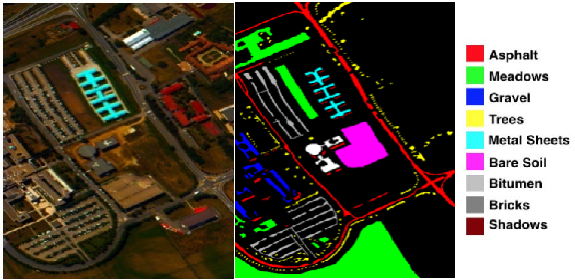}
\caption{False color composite and ground truth labels of Pavia University dataset}
\label{fig:pavia_university}
\end{figure}

The Pavia University dataset was acquired in 2001 by the ROSIS imaging spectrometer over the Pavia region in northern Italy. The dataset contains images of size $610\times340$ pixels with a spatial resolution of 1.3 meters, comprising 115 spectral bands in the wavelength range of 0.43--0.86$\mu$m. For our experiments, we removed 12 bands containing strong noise and water vapor absorption, retaining 103 bands for analysis. The dataset includes 9 land cover categories such as roads, trees, and roofs. Figure~\ref{fig:pavia_university} shows the spatial distribution of different classes.

\subsubsection{Kennedy Space Center Dataset}
The KSC dataset was collected on March 23, 1996 by the AVIRIS imaging spectrometer over the Kennedy Space Center in Florida. The AVIRIS sensor captured 224 spectral bands with 10 nm width, centered at wavelengths from 400 to 2500 nm. Acquired from an altitude of approximately 20 km, the dataset has a spatial resolution of 18 meters. After removing bands affected by water absorption and low signal-to-noise ratio (SNR), we used 176 bands for analysis, which represent 13 defined land cover categories.

\subsection{Experimental Analysis}
STNet was trained for 80 epochs on all three datasets using the Adam optimizer. The experiments were conducted on a platform with four 80GB A100 GPUs. For analysis, we employed the stnet-base architecture with three stages, where each stage contained 4, 6, and 8 dense blocks respectively. The growth rates were set to 8, 16, and 32, with 4 heads. The 3×3 group convolution used 4 groups, gate factor was 0.25, and the compression ratio was 16.

\subsubsection{Data Partitioning Ratio}
For hyperspectral data with limited samples, the training set ratio significantly impacts model performance. To systematically evaluate the sensitivity of data partitioning strategies, we compared the model's generalization performance under different Train/Validation/Test ratios. Experiments show that with limited training samples, a 6:1:3 ratio effectively balances learning capability and evaluation reliability on Indian Pines - this configuration allocates 60\% samples for training, 10\% for validation (enabling early stopping to prevent overfitting), and 30\% for statistically significant testing. STNet adopted 4:1:5 ratio on Pavia University and 5:1:4 ratio on KSC datasets, with 11×11 neighboring pixel blocks to balance local feature extraction and spatial context integrity.

\begin{table}[h]
\begin{minipage}{\textwidth}
\centering
\makeatletter
\def\@makecaption#1#2{%
    \vskip\abovecaptionskip
    \centering 
    \small #1: #2\par
    \vskip\belowcaptionskip
}
\makeatother
\caption{OA, AA and Kappa metrics for different training set ratios on the Indian Pines dataset}
\begin{tabular}{cccc}
\toprule
Training Ratio & OA & AA & Kappa \\
\midrule
2:1:7 & 96.92 & 96.34 & 96.48 \\
3:1:6 & 98.58 & 98.39 & 98.39 \\
4:1:5 & 99.14 & 98.73 & 99.03 \\
5:1:4 & 99.46 & 99.30 & 99.39 \\
6:1:3 & 99.67 & 99.57 & 99.63 \\
\bottomrule
\end{tabular}
\label{tab:indian_ratios}
    \end{minipage}

 \vspace*{10pt} 

\begin{minipage}{\textwidth}
\centering
\makeatletter
\def\@makecaption#1#2{%
    \vskip\abovecaptionskip
    \centering 
    \small #1: #2\par
    \vskip\belowcaptionskip
}
\makeatother
\caption{OA, AA and Kappa metrics for different training set ratios on the Pavia University dataset}
\begin{tabular}{cccc}
\toprule
Training Ratio & OA & AA & Kappa \\
\midrule
2:1:7 & 99.50 & 99.37 & 99.34 \\
3:1:6 & 97.53 & 97.39 & 96.73 \\
4:1:5 & 99.91 & 99.76 & 99.89 \\
5:1:4 & 99.93 & 99.93 & 99.92 \\
6:1:3 & 99.95 & 99.92 & 99.93 \\
\bottomrule
\end{tabular}
\label{tab:pavia_ratios}
        \end{minipage}

 \vspace*{10pt} 
 
\begin{minipage}{\textwidth} 
\centering
\makeatletter
\def\@makecaption#1#2{%
    \vskip\abovecaptionskip
    \centering 
    \small #1: #2\par
    \vskip\belowcaptionskip
}
\caption{OA, AA and Kappa metrics for different training set ratios on the KSC dataset}
\begin{tabular}{cccc}
\toprule
Training Ratio & OA & AA & Kappa \\
\midrule
2:1:7 & 92.92 & 90.88 & 92.11 \\
3:1:6 & 98.97 & 98.25 & 98.86 \\
4:1:5 & 99.65 & 99.47 & 99.62 \\
5:1:4 & 99.86 & 99.69 & 99.84 \\
6:1:3 & 99.61 & 99.55 & 99.57 \\
\bottomrule
\end{tabular}
\label{tab:ksc_ratios}
        \end{minipage}
\end{table}

\subsubsection{Neighboring Pixel Blocks}
The network performs edge padding on the input $145\times145\times103$ image (using Indian Pines as an example), transforming it into a $155\times155\times103$ image. On this padded image, it sequentially selects adjacent pixel blocks of size $M\times N\times L$, where $M\times N$ represents the spatial sampling size and $L$ is the full spectral dimension. Large original images are unfavorable for convolutional feature extraction, leading to slower processing speeds, temporary memory spikes, and higher hardware requirements. Therefore, we adopt adjacent pixel block processing. The block size is a crucial hyperparameter - too small blocks may result in insufficient receptive fields for convolutional feature extraction, leading to poor local performance. As shown in Tables 4-6, on the Indian Pines dataset, accuracy shows significant improvement when block size increases from 7 to 17. However, the accuracy gain diminishes with larger blocks, showing a clear threshold effect. At block size 17, accuracy even decreases slightly, a phenomenon also observed in Pavia University and KSC datasets. Consequently, we select block size 15 for Indian Pines and 17 for Pavia University and KSC datasets.

\begin{table*}[h]
\begin{minipage}{\textwidth}
\centering
\makeatletter
\def\@makecaption#1#2{%
    \vskip\abovecaptionskip
    \centering 
    \small #1: #2\par
    \vskip\belowcaptionskip
}
\makeatother
\caption{OA, AA and Kappa metrics for different block sizes on Indian Pines}
\begin{tabular}{cccc}
\toprule
Block Size (M=N) & OA & AA & Kappa \\
\midrule
7 & 99.54 & 99.50 & 99.48 \\
9 & 99.71 & 99.53 & 99.67 \\
11 & 99.67 & 99.57 & 99.63 \\
13 & 99.77 & 99.66 & 99.74 \\
15 & - & - & - \\
17 & - & - & - \\
\hline
\end{tabular}
        \end{minipage}

 \vspace*{10pt} 

\begin{minipage}{\textwidth}
\centering
\makeatletter
\def\@makecaption#1#2{%
    \vskip\abovecaptionskip
    \centering 
    \small #1: #2\par
    \vskip\belowcaptionskip
}
\makeatother
\caption{OA, AA and Kappa metrics for different block sizes on Pavia University}
\begin{tabular}{cccc}
\toprule
Block Size (M=N) & OA & AA & Kappa \\
\midrule
7 & 99.89 & 99.88 & 99.85 \\
9 & 99.92 & 99.88 & 99.90 \\
11 & 99.95 & 99.92 & 99.93 \\
13 & 99.98 & 99.97 & 99.97 \\
15 & 99.97 & 99.96 & 99.95 \\
17 & 1 & 1 & 1 \\
\hline
\end{tabular}
\end{minipage}

 \vspace*{10pt} 
 
\begin{minipage}{\textwidth}
\centering
\makeatletter
\def\@makecaption#1#2{%
    \vskip\abovecaptionskip
    \centering 
    \small #1: #2\par
    \vskip\belowcaptionskip
}
\makeatother
\caption{OA, AA and Kappa metrics for different block sizes on KSC}
\begin{tabular}{cccc}
\toprule
Block Size (M=N) & OA & AA & Kappa \\
\midrule
7 & 98.56 & 97.70 & 98.39 \\
9 & 99.66 & 99.42 & 99.63 \\
11 & 99.86 & 99.69 & 99.84 \\
13 & 99.81 & 99.64 & 99.79 \\
15 & 99.95 & 99.92 & 99.95 \\
17 & - & - & - \\
\hline
\end{tabular}
        \end{minipage}
\end{table*}

\subsection{Network Parameters}
We categorize STNet into base and large types. Table 7 shows OA, AA, and Kappa results on Indian Pines dataset. While the large model with more parameters shows improved accuracy, the gain is not substantial compared to the rapid parameter increase in dense blocks, suggesting STNet-base already possesses excellent generalization and classification capabilities.

\begin{table*}[!ht]
\centering
\makeatletter
\def\@makecaption#1#2{%
    \vskip\abovecaptionskip
    \centering 
    \small #1: #2\par
    \vskip\belowcaptionskip
}
\makeatother
\caption{STNet model configurations and performance metrics}
\begin{tabular}{cccccc}
\toprule
Model & Stages/Dense Block & Growth Rate & OA & AA & Kappa \\
\midrule
STNet-base & 4,6,8 & 8,16,32 & 99.77 & 99.66 & 99.74 \\
STNet-large & 14,14,14 & 8,16,32 & 99.87 & 99.67 & 99.88 \\
\bottomrule
\end{tabular}
\label{tab:params}
\end{table*}

\subsection{Experimental Results}
On Indian Pines, Pavia University, and KSC datasets, STNet uses input sizes of $17\times17\times200$, $17\times17\times103$, and $17\times17\times176$ respectively. We compare STNet-base with SSRN, 3D-CNN, 3D-SE-DenseNet, Spectralformer, lgcnet, and dgcnet (Tables 8-9). STNet variants consistently achieve leading accuracy. Figure 9 shows training/validation loss and accuracy curves, demonstrating rapid convergence and stable accuracy improvement.

\begin{table*}[!ht]
\centering
\makeatletter
\def\@makecaption#1#2{%
    \vskip\abovecaptionskip
    \centering 
    \small #1: #2\par
    \vskip\belowcaptionskip
}
\makeatother
\caption{Classification accuracy comparison (\%) on Indian Pines dataset}
\label{tab:indian_results}
\resizebox{\textwidth}{!}{
\begin{tabular}{@{}lccccccc@{}}
\toprule
Class & SSRN & 3D-CNN & 3D-SE-DenseNet & DGCNet & Hit & Spectralformer & STNet \\
\midrule
1 & 100 & 96.88 & 95.87 & 100 & Hit & 70.52 & 100 \\
2 & 99.85 & 98.02 & 98.82 & 99.47 & 94.25 & 81.89 & 99.52 \\
3 & 99.83 & 97.74 & 99.12 & 99.51 & 92.68 & 91.30 & 100 \\
4 & 100 & 96.89 & 94.83 & 97.65 & 78.55 & 95.53 & 100 \\
5 & 99.78 & 99.12 & 99.86 & 100 & 86.73 & 85.51 & 100 \\
6 & 99.81 & 99.41 & 99.33 & 99.88 & 85.33 & 99.32 & 100 \\
7 & 100 & 88.89 & 97.37 & 100 & 98.32 & 81.81 & 100 \\
8 & 100 & 100 & 100 & 100 & 92.00 & 75.48 & 100 \\
9 & 0 & 100 & 100 & 100 & 94.63 & 73.76 & 100 \\
10 & 100 & 100 & 99.48 & 98.85 & 64.86 & 98.77 & 99.34 \\
11 & 99.62 & 99.33 & 98.95 & 99.72 & 89.48 & 93.17 & 99.86 \\
12 & 99.17 & 97.67 & 95.75 & 99.56 & 94.40 & 78.48 & 99.45 \\
13 & 100 & 99.64 & 99.28 & 100 & 89.32 & 100 & 100 \\
14 & 98.87 & 99.65 & 99.55 & 99.87 & 99.46 & 79.49 & 100 \\
15 & 100 & 96.34 & 98.70 & 100 & 97.23 & 100 & 100 \\
16 & 98.51 & 97.92 & 96.51 & 98.30 & 68.71 & 100 & 96.43 \\
\midrule
OA & 99.62±0.00 & 98.23±0.12 & 98.84±0.18 & 99.58 & 91.67 & 81.76 & 99.77 \\
AA & 93.46±0.50 & 98.80±0.11 & 98.42±0.56 & 99.55 & 87.85 & 87.81 & 99.66 \\
K & 99.57±0.00 & 97.96±0.53 & 98.60±0.16 & 99.53 & 88.12 & 79.19 & 99.74 \\
\bottomrule
\end{tabular}
}
\end{table*}

\begin{figure*}[!ht]
\centering
\includegraphics[width=0.8\linewidth]{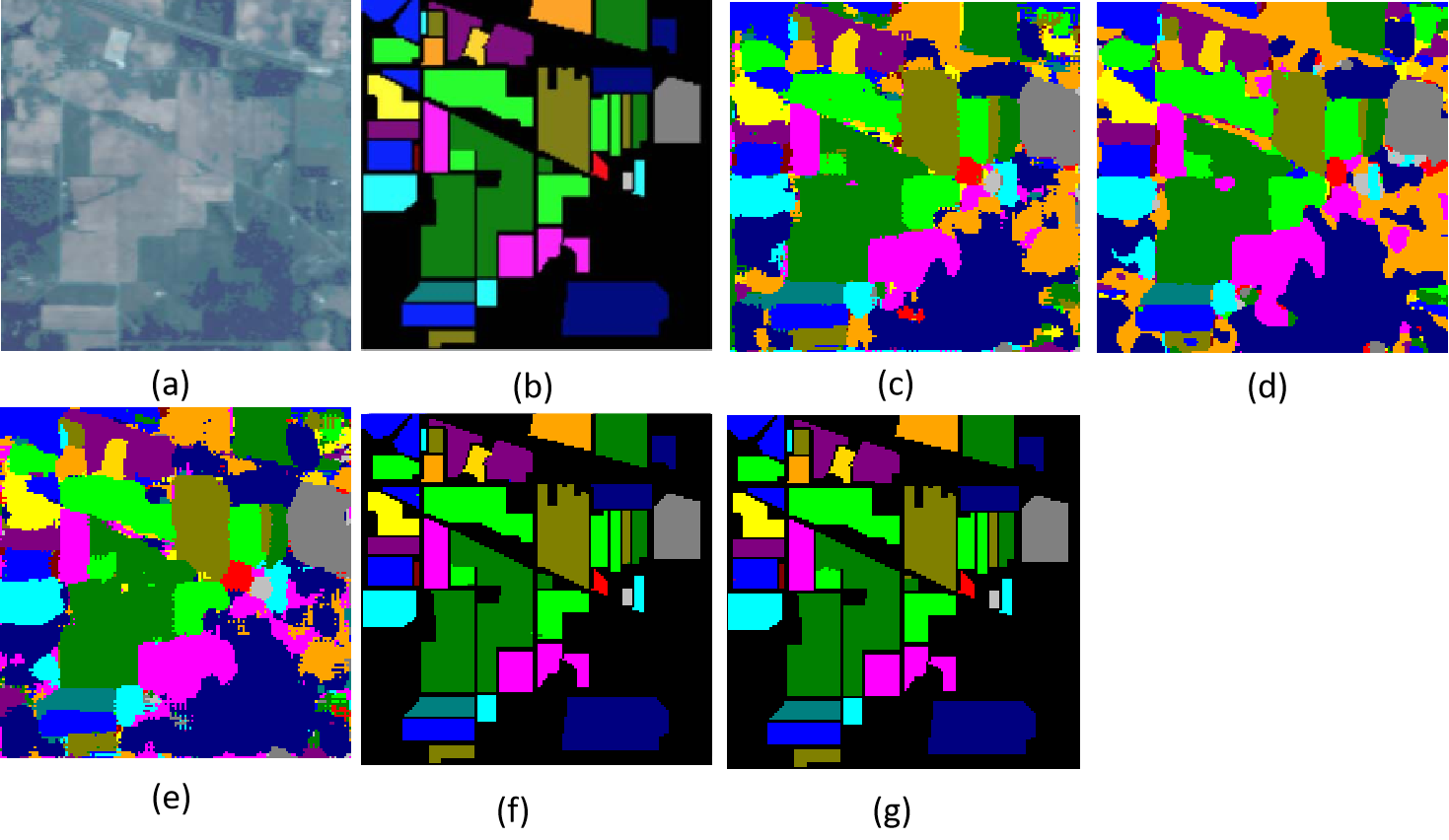}
\caption{Classification results comparison for Indian Pines dataset: (a) False color image, (b) Ground-truth labels, (c)-(g) Classification results of SSRN, 3D-CNN, 3D-SE-DenseNet-BC, DGCNet, and STNet}
\label{fig:classification}
\end{figure*}

\begin{figure*}[ht]
\centering
\includegraphics[width=0.8\linewidth]{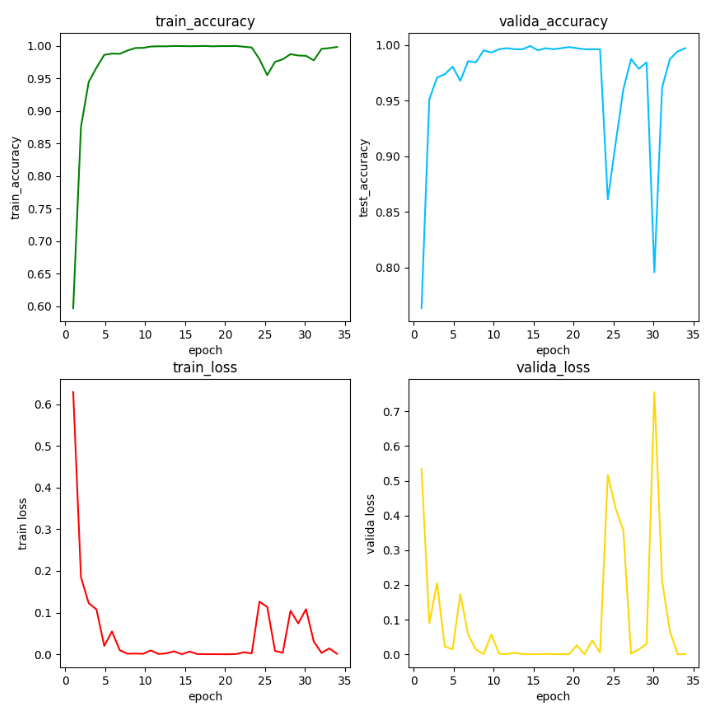}
\caption{Training and validation curves of STNet showing loss and accuracy evolution}
\label{fig:training_curve}
\end{figure*}

\begin{table*}[!ht]
\centering
\makeatletter
\def\@makecaption#1#2{%
    \vskip\abovecaptionskip
    \centering 
    \small #1: #2\par
    \vskip\belowcaptionskip
}
\makeatother
\caption{Classification accuracy (\%) comparison on Pavia University dataset}
\label{tab:pavia_results}
\resizebox{\textwidth}{!}{
\begin{tabular}{@{}lcccccc@{}}
\toprule
Class & SSRN & 3D-CNN & 3D-SE-DenseNet & Hit & Spectralformer & STNet \\
\midrule
1 & 89.93 & 99.96 & 99.32 & 96.19 & 82.73 & 100 \\
2 & 86.48 & 99.99 & 99.87 & 92.79 & 94.03 & 100 \\
3 & 99.95 & 99.64 & 96.76 & 93.21 & 73.66 & 100 \\
4 & 95.78 & 99.83 & 99.23 & 97.33 & 93.75 & 100 \\
5 & 97.69 & 99.81 & 99.64 & 99.96 & 99.28 & 100 \\
6 & 95.44 & 99.98 & 99.80 & 99.91 & 90.75 & 100 \\
7 & 84.40 & 97.97 & 99.47 & 98.22 & 87.56 & 100 \\
8 & 100 & 99.56 & 99.32 & 99.15 & 95.81 & 100 \\
9 & 87.24 & 100 & 100 & 99.77 & 94.21 & 100 \\
\midrule
OA & 92.99±0.39 & 99.79±0.01 & 99.48±0.02 & 92.00 & 91.07 & 100 \\
AA & 87.21±0.25 & 99.75±0.15 & 99.16±0.37 & 93.24 & 90.20 & 100 \\
K & 90.58±0.18 & 99.87±0.27 & 99.31±0.03 & 89.77 & 88.05 & 100 \\
\bottomrule
\end{tabular}
}
\end{table*}

\section{Conclusion}
This paper proposes a novel STNet architecture, whose core innovation lies in its SpatioTemporalTransformer module. This module replaces the traditional 3D convolution's mixed, fixed processing of spatial-spectral information through explicit spatial-spectral attention decoupling mechanisms (spatial\_attn, temporal\_attn) combined with adaptive attention fusion gating (gate), enabling more precise separate capture and intelligent integration of complex local spatial structures and long-range spectral dependencies in hyperspectral data. Simultaneously, the gated feed-forward network (GFFN, gate\_ffn) enhances information screening capability during feature transformation, while the learnable and interpolatable 3D positional encoding (pos\_embed) improves model adaptability to different inputs. This design demonstrates powerful feature representation capability, particularly effective in reducing overfitting risks in small-sample and high-noise scenarios. STNet optimizes representation of joint spatial-spectral information through focused attention on key features while suppressing redundant information, integrated within the 3D-DenseNet architecture to leverage its feature reuse advantages, providing an efficient and robust solution for hyperspectral image classification that successfully addresses challenges posed by sparse ground object distribution and spectral redundancy.

\section*{Data Availability Statement}
The datasets used in this study are publicly available and widely used benchmark datasets in the hyperspectral image analysis community.

{\small
\bibliographystyle{template}
\bibliography{template}
}

\end{document}